\documentclass{bmvc2k}

\usepackage{hyperref}
\usepackage{gensymb,graphics,subfigure,inputenc}
\usepackage[linesnumbered,algo2e,boxed]{algorithm2e}
\usepackage[figuresright]{rotating}
\usepackage{textcomp,subfigure,multirow,upgreek}
\usepackage{booktabs}
\usepackage{mdwlist}
\usepackage{amsmath}
\usepackage{amssymb}

\usepackage{ragged2e}
\usepackage{fontawesome}
\graphicspath{{Figures/}}

\usepackage{caption}
\usepackage{enumitem}
\setitemize{noitemsep,topsep=0pt,parsep=0pt,partopsep=0pt}

\title{Bipartite Graph Reasoning GANs for Person Image Generation}

\addauthor{Hao Tang}{hao.tang@unitn.it}{12}
\addauthor{Song Bai}{songbai.site@gmail.com}{2}
\addauthor{Philip H.S. Torr}{philip.torr@eng.ox.ac.uk}{2}
\addauthor{Nicu Sebe}{sebe@disi.unitn.it}{13}

\addinstitution{
DISI \\
University of Trento
}
\addinstitution{
Department of Engineering Science \\
University of Oxford\\
}
\addinstitution{
Huawei Research Ireland
}

\runninghead{Hao Tang et al.}{Bipartite Graph Reasoning GANs}

\begin{document}

\maketitle

\begin{abstract}
We present a novel Bipartite Graph Reasoning GAN (BiGraphGAN) for the challenging person image generation task.
The proposed graph generator mainly consists of two novel blocks that aim to model the pose-to-pose and pose-to-image relations, respectively. 
Specifically, the proposed Bipartite Graph Reasoning (BGR) block aims to reason the crossing long-range relations between the source pose and the target pose in a bipartite graph, which mitigates some challenges caused by pose deformation.
Moreover, we propose a new Interaction-and-Aggregation (IA) block to effectively update and enhance the feature representation capability of both person's shape and appearance in an interactive way.
Experiments on two challenging and public datasets,~\emph{i.e.},~Market-1501 and DeepFashion, show the effectiveness of the proposed BiGraphGAN in terms of objective quantitative scores and subjective visual realness.
The source code and trained models are available at
\url{https://github.com/Ha0Tang/BiGraphGAN}.
\end{abstract}
\section{Introduction}
\label{sec:intro}
\begin{figure}[!t]
	\centering
	\includegraphics[width=1\linewidth]{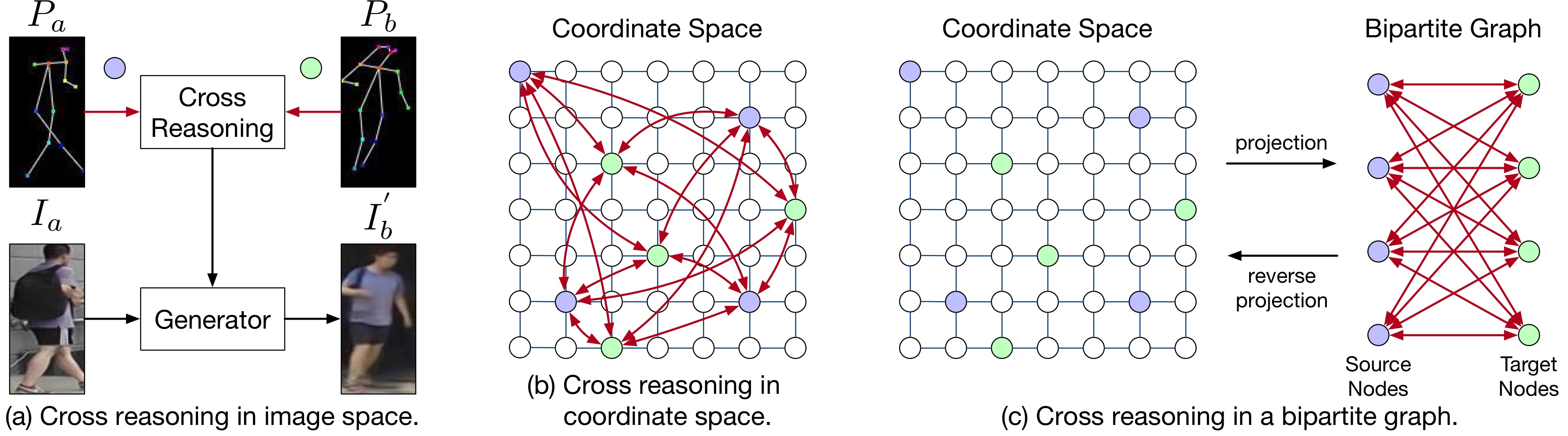}
	\caption{Illustration of our motivation. We propose a novel BiGraphGAN (Fig.~(c)) for capturing crossing long-range relations between the source pose $P_a$ and the target pose $P_b$ in a bipartite graph. The node features from both source and target poses in the coordinate space are projected into the nodes in a bipartite graph, thereby forming a fully-connected bipartite graph. After cross-reasoning the graph, the node features are projected back to the original coordinate space for further processing.}
	\label{fig:motivation}
\end{figure}

\begin{figure}[!t]
	\centering
	\includegraphics[width=1\linewidth]{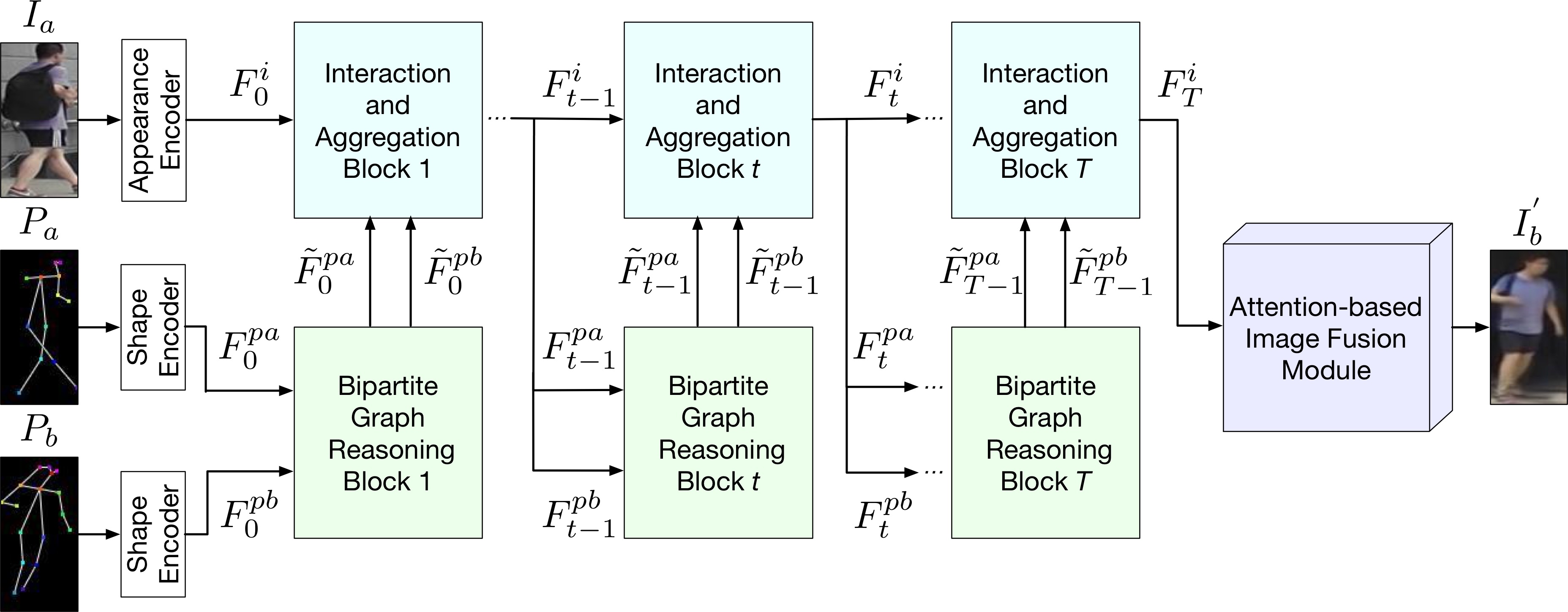}
	\caption{Overview of the proposed graph generator, which consists of a sequence of Bipartite Graph Reasoning (BGR) blocks, a sequence of Interaction-and-Aggregation (IA) blocks and an Attention-based Image Fusion (AIF) module. BGR blocks aim to reason the crossing long-range relations between the source pose and the target pose in a bipartite graph. IA blocks aim to interactively update person's appearance and shape feature representations. AIF module aims to selectively generate the final result via an attention network.
	The symbols $F^i{=}\{F^i_j\}_{j=0}^T$, $F^{pa}{=}\{F^{pa}_j\}_{j=0}^{T{-}1}$, $F^{pb}{=}\{F^{pb}_j\}_{j=0}^{T{-}1}$, $\tilde{F}^{pa}{=}\{\tilde{F}^{pa}_j\}_{j=0}^{T{-}1}$, and $\tilde{F}^{pb}{=}\{\tilde{F}^{pb}_j\}_{j=0}^{T{-}1}$ denote the appearance codes, the source shape codes, the target shape codes, the updated source shape codes, and the updated target shape codes, respectively.
	}
	\label{fig:method}
\end{figure}

In this paper, we mainly focus on translating a person image from one pose to another as depicted in Fig.~\ref{fig:motivation} and~\ref{fig:method}.
Existing person image generation methods such as \cite{ma2017pose,ma2018disentangled,siarohin2018deformable,tang2019cycle,albahar2019guided,esser2018variational,zhu2019progressive,chan2019everybody,balakrishnan2018synthesizing,zanfir2018human,liang2019pcgan,liu2019liquid} always rely on building convolution layers. 
Due to the physical design of convolutional filters, convolution operations can only model local relations.
To capture global relations, existing methods such as \cite{zhu2019progressive,tang2019cycle} inefficiently stack multiple convolution layers to enlarge the receptive fields to cover all the body joints from both the source pose and the target pose.
However, none of the above-mentioned methods explicitly consider modeling the cross relations between the source pose and the target pose.

In this paper, we propose a novel Bipartite Graph Reasoning GAN (BiGraphGAN), which mainly consists of two novel blocks, \emph{i.e.}, Bipartite Graph Reasoning (BGR) block and Interaction-and-Aggregation (IA) block.
The BGR block aims to efficiently capture the crossing long-range relations between the source pose and the target pose in a bipartite graph (see Fig.~\ref{fig:motivation}). 
Specifically, the BGR block first projects both the source pose feature and the target pose feature in the original coordinate space onto a bipartite graph.
Next, both source and target pose features are represented by a set of nodes to form a fully-connected bipartite graph, on which crossing long-range relation reasoning is performed by Graph Convolution Networks (GCNs). 
To the best of our knowledge, we are the first to explore GCNs to model the crossing long-range relations for solving the challenging person image generation task.
After reasoning, we project the node features back to the original coordinate space for further processing. 

Also, the proposed IA block is proposed to effectively and interactively enhance person's shape and appearance features.
We also introduce an Attention-based Image Fusion (AIF) module to selectively generate the final result using an attention network.
Qualitative and quantitative experiments on two challenging datasets, \emph{i.e.}, Market-1501 \cite{zheng2015scalable} and DeepFashion \cite{liu2016deepfashion}, demonstrate that the proposed BiGraphGAN generates better person images than several state-of-the-art methods, \emph{i.e.}, PG2~\cite{ma2017pose}, DPIG~\cite{ma2018disentangled}, Deform~\cite{siarohin2018deformable}, C2GAN~\cite{tang2019cycle}, BTF~\cite{albahar2019guided}, VUnet~\cite{esser2018variational} and PATN~\cite{zhu2019progressive}.

The contributions of this paper are summarized as follows,
\begin{itemize}[leftmargin=*]
	\item We propose a novel Bipartite Graph Reasoning GAN (BiGraphGAN) for person image generation. The proposed BiGraphGAN aims to progressively reason the pose-to-pose and pose-to-image relations via two novel proposed blocks.
	\item We propose a novel Bipartite Graph Reasoning (BGR) block to effectively reason the crossing long-range relations between the source pose and the target pose in a bipartite graph by using Graph Convolutional Networks (GCNs).
	Moreover, we present a new Interaction-and-Aggregation (IA) block to interactively enhance both person's appearance and shape feature representations.
	\item Extensive experiments on two challenging datasets, \emph{i.e.}, Market-1501 \cite{zheng2015scalable} and DeepFashion \cite{liu2016deepfashion}, demonstrate the effectiveness of the proposed BiGraphGAN and show significantly better performance compared with state-of-the-art approaches.
\end{itemize}
\section{Related Work}
\noindent \textbf{Generative Adversarial Networks (GANs)} \cite{goodfellow2014generative} have shown the potential to generate realistic images \cite{shaham2019singan,karras2019style,brock2019large}.
For instance, Shaham \emph{et al.} propose an unconditional SinGAN~\cite{shaham2019singan} which can be learned from a single image.
Moreover, to generate user-defined images, Conditional GAN (CGAN) \cite{mirza2014conditional} has been proposed recently.
A CGAN always consists of a vanilla GAN and external guide information such as class labels \cite{wu2019relgan,choi2018stargan,zhang2018sparsely}, segmentation maps \cite{tang2019multi,park2019semantic,tang2020local,liu2020exocentric}, attention maps \cite{kim2019u,tang2019attention,mejjati2018unsupervised}, and human skeleton \cite{albahar2019guided,balakrishnan2018synthesizing,zhu2019progressive,tang2018gesturegan,tang2020xinggan}.
In this work, we mainly focus on the challenging person image generation task, which aims to transfer a person image from one pose to another one. 

\noindent \textbf{Person Image Generation} is a challenging task due to the pose deformation between the source image and the target image.
Modeling the long-range relations between the source pose and the target pose is the key to solving this challenging task.
However, existing methods such as \cite{ma2017pose,ma2018disentangled,balakrishnan2018synthesizing,siarohin2018deformable,tang2019cycle,albahar2019guided,esser2018variational,zhu2019progressive,chan2019everybody,zanfir2018human,liang2019pcgan,liu2019liquid} built through the stacking of convolutional layers, which can only leverage the relations between the source pose and the target pose locally.
For instance, Zhu \emph{et al.}~\cite{zhu2019progressive} propose a Pose-Attentional Transfer Block (PATB), in which the source and target poses are simply concatenated and then fed into an encoder to capture their dependencies.

Unlike existing methods for modeling the relations between the source and target poses in a localized manner, we show that the proposed Bipartite Graph Reasoning (BGR) block can bring considerable performance improvements in the global view.

\noindent \textbf{Graph-Based Reasoning.} Graph-based approaches have shown to be an efficient way to reason relation in many computer vision tasks such as semi-supervised classification \cite{kipf2017semi}, video recognition \cite{wang2018videos}, crowd counting~\cite{chen2020relevant}, action recognition \cite{yan2018spatial,peng2020mix} and semantic segmentation \cite{chen2019graph,zhang2019dual}.

Compared to these graph-based reasoning methods which model the long-range relations within the same feature map to incorporate global information, we focus on developing a novel BiGraphGAN framework that reasons and models the crossing long-range relations between different features of the source pose and target pose in a bipartite graph.
Then the crossing relations are further used to guide the image generation process (see Fig.~\ref{fig:motivation}). 
This idea has not been investigated in existing GAN-based image translation methods.

\section{Bipartite Graph Reasoning GANs}

\begin{figure}[!t]
	\centering
	\includegraphics[width=1\linewidth]{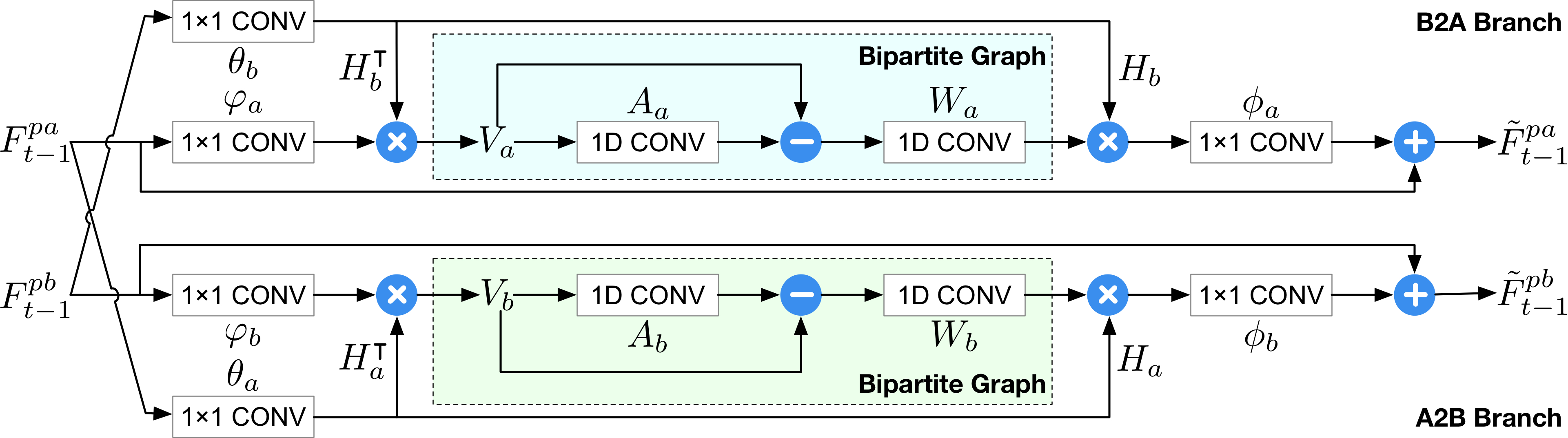}
	\caption{Illustration of the proposed Bipartite Graph Reasoning (BGR) Block $t$, which consists of two branches, \emph{i.e.}, B2A and A2B. Each of them aims to model cross-contextual information between shape features $F_{t-1}^{pa}$ and $F_{t-1}^{pb}$ in a bipartite graph via Graph Convolutional Networks (GCNs).}
	\label{fig:blocks}
\end{figure}

We start by introducing the details of the proposed Bipartite Graph Reasoning GAN (BiGraphGAN), which consists of a graph generator $G$ and two discriminators (\emph{i.e.}, appearance discriminator $D_a$ and shape discriminator $D_s$).
An illustration of the proposed graph generator $G$ is shown in Fig.~\ref{fig:method}, which mainly contains three parts, \emph{i.e.}, a sequence of Bipartite Graph Reasoning (BGR) blocks modeling the crossing long-range relations between the source pose $P_a$ and the target pose $P_b$, a sequence of Interaction-and-Aggregation (IA) blocks interactively enhancing both person's shape and appearance feature representations, and an Attention-based Image Fusion (AIF) module attentively generating the final result~$I_b^{'}$.
In the following, we first present the proposed blocks and then introduce the optimization objective and implementation details of the proposed BiGraphGAN.

Fig.~\ref{fig:method} shows the proposed graph generator $G$, whose inputs are the source image $I_a$, the source pose $P_a$ and the target pose $P_b$.
The generator $G$ aims to transfer the pose of the person in the source image $I_a$ from the source pose $P_a$ to the target pose $P_b$, generating the desired image $I_b^{'}$.
Firstly, $I_a$, $P_a$ and $P_b$ are separately fed into three encoders to obtain the appearance code $F_0^i$, the source shape code $F_0^{pa}$ and the target shape code $F_0^{pb}$.
Note that we used the same shape encoder to learn both $P_a$ and $P_b$, \emph{i.e.}, the two shape encoders for learning the two different poses are sharing the weights.

\subsection{Pose-to-Pose Bipartite Graph Reasoning}
The proposed Bipartite Graph Reasoning (BGR) block aims to reason the crossing long-range relations between the source pose and the target pose in a bipartite graph.
All BGR blocks have an identical structure as illustrated in Fig.~\ref{fig:method}. 
Consider the $t$-th block given in Fig.~\ref{fig:blocks}, whose inputs are the source shape code $F_{t-1}^{pa}$ and the target shape code $F_{t-1}^{pb}$.
The BGR block aims to reason these two codes in a bipartite graph via Graph Convolutional Networks (GCNs) and outputs new shape codes.
The proposed BGR block contains two symmetrical branches (\emph{i.e.}, B2A branch and A2B branch) because a bipartite graph is a bidirectional graph. As shown in Fig.~\ref{fig:motivation}(c), each node in the source nodes connects all the target nodes; at the same time, each node in the target nodes connects all the source nodes.
In the following, we mainly describe the detailed modeling process of the B2A branch, and another A2B branch is similar to this.

\noindent \textbf{From Coordinate Space to Bipartite-Graph Space.}
Firstly, we reduce the dimension of the source shape code $F_{t-1}^{pa}$ with function $\varphi_a(F_{t-1}^{pa}) {\in} \mathbb{R}^{C \times D_a}$, where $C$ is the number of feature map channels, $D_a$ is the number of nodes of $F_{t-1}^{pa}$.
Then we reduce the dimension of the target shape code $F_{t-1}^{pb}$ with function $\theta_b(F_{t-1}^{pb}) {=} H_b^\intercal {\in} \mathbb{R}^{D_b  \times C}$, where $D_b$ is the number of nodes of $F_{t-1}^{pb}$.
Next, we project $F_{t-1}^{pa}$ to a new feature $V_a$ in a bipartite graph using the projection function $H_b^T$. 
Therefore we have,
\begin{equation}
\begin{aligned}
V_a = H_b^\intercal \varphi_a(F_{t-1}^{pa}) = \theta_b(F_{t-1}^{pb}) \varphi_a(F_{t-1}^{pa}),
\end{aligned}
\end{equation}
where both functions $\theta_b(\cdot)$ and $\varphi_a(\cdot)$ are implemented using $1{\times1}$ convolutional layer. 
This results in a new feature $V_a {\in} \mathbb{R}^{D_b \times D_a}$ in the bipartite graph, which represents the crossing relations between the nodes of the target pose $F_{t-1}^{pb}$ and the source pose $F_{t-1}^{pa}$ (see Fig.~\ref{fig:motivation}(c)).

\noindent \textbf{Cross Reasoning with Graph Convolution.}
After projection, we build a fully-connected bipartite graph with adjacency matrix $A_a {\in} \mathbb{R}^{D_b \times D_b}$. 
We then use a graph convolution to reason the crossing long-range relations between the nodes from both source and target poses, which can be formulated as,
\begin{equation}
\begin{aligned}
M_a = ({\rm I} - A_a) V_a W_a,
\end{aligned}
\end{equation}
where $W_a {\in} \mathbb{R}^{D_a \times D_a}$ denotes the trainable edge weights.
We follow \cite{chen2019graph,zhang2019dual} and use Laplacian smoothing  \cite{chen2019graph,li2018deeper} to propagate the node features over the bipartite graph. 
The identity matrix~${\rm I}$ can be viewed as a residual sum connection to alleviate optimization difficulties. 
Note that we randomly initialize both adjacency matrix $A_a$ and the weights $W_a$, and then train both by gradient descent in an end-to-end manner.

\noindent \textbf{From Bipartite-Graph Space to Coordinate Space.}
After the cross-reasoning, the updated new feature $M_a$ is mapped back to the original coordinate space for further processing.
Next, we add the result to the original source shape code $F_{t-1}^{pa}$ to form a residual connection \cite{he2016deep}.
This process can be expressed as,
\begin{equation}
\begin{aligned}
\tilde{F}_{t-1}^{pa} = \phi_a(H_b M_a) + F_{t-1}^{pa},
\end{aligned}
\end{equation}
where we reuse the projection matrix $H_b$ and perform a linear projection
$\phi_a(\cdot)$ to project $M_a$ back to the original coordinate space. 
Therefore, we obtain the new source feature $\tilde{F}_{t-1}^{pa}$, which has the same dimension with the original one $F_{t-1}^{pa}$.

Similarly, the A2B branch outputs the new target shape feature $\tilde{F}_{t-1}^{pb}$. 
Note that the idea of the proposed BGR block is inspired by the GloRe unit proposed by \cite{chen2019graph}. 
The main difference is that the GloRe unit reasons the relations within the same feature map via a standard graph, but the proposed BGR block reasons the crossing relations between feature maps of different poses using a bipartite graph.

\subsection{Pose-to-Image Interaction and Aggregation}
As shown in Fig.~\ref{fig:method}, the proposed Interaction-and-Aggregation (IA) block receives the appearance code $F_{t-1}^i$, the new source shape code $\tilde{F}_{t-1}^{pa}$ and the new target shape code $\tilde{F}_{t-1}^{pb}$ as inputs.
IA block aims to simultaneously and interactively enhance $F_{t}^i$, $F_{t}^{pa}$ and $F_{t}^{pb}$.
Specifically, both shape codes firstly concatenated and fed into two convolutional layers to produce the attention map $A_p$.
Mathematically,
\begin{equation}
\begin{aligned}
A_p = \sigma({\rm Conv}({\rm Concat}(\tilde{F}_{t-1}^{pa}, \tilde{F}_{t-1}^{pb}))),
\end{aligned}
\end{equation}
where $\sigma(\cdot)$ denotes the element-wise Sigmoid function. 

\noindent \textbf{Appearance Code Enhance.} After obtaining $A_p$, the appearance $F_{t-1}^i$ is enhanced by,
\begin{equation}
\begin{aligned}
F_t^i = A_p \otimes  F_{t-1}^i  + F_{t-1}^i,
\end{aligned}
\end{equation}
where $\otimes$ denotes element-wise product. By multiplying with the attention map $A_p$, the new appearance code $F_t^i$ at certain locations can be either preserved or suppressed. 

\noindent \textbf{Shape Code Enhance.}
Next, we concatenate $F_t^i $, $F_{t-1}^{pa}$ and $F_{t-1}^{pb}$, and go through two convolutional layers to obtain the updated shape code $F_t^{pa}$ and $F_t^{pb}$ by splitting the result along the channel axis.
This process can be performed by,
\begin{equation}
\begin{aligned}
F_{t}^{pa}, F_{t}^{pb} = {\rm Conv} ({\rm Concat}(F_t^i, \tilde{F}_{t-1}^{pa}, \tilde{F}_{t-1}^{pb})).
\end{aligned}
\end{equation}
In this way, both new shape codes $F_{t}^{pa}$ and $F_{t}^{pb}$ can synchronize the changes caused by the new appearance code $F_t^i$. 

\subsection{Attention-Based Image Fusion}
At the $T$-th IA block, we obtain the final appearance code $F_T^{i}$.
We then feed $F_T^{i}$ to an image decoder to generate the intermediate result $\tilde{I}_b$.
At the same time, we feed $F_T^{i}$ to an attention decoder to produce the attention mask $A_i$.

The attention encoder consists of several deconvolutional layers and a Sigmoid activation layer. Thus, the attention encoder aims to generate a one-channel attention mask $A_i$, in which each pixel value is between 0 to 1.
The attention mask $A_i$ aims to selectively pick useful content from both the input image $I_a$ and the intermediate result $\tilde{I}_b$ for generating the final result~$I_b^{'}$. This process can be expressed as,
\begin{equation}
\begin{aligned}
I_b^{'} = I_a \otimes  A_i + \tilde{I}_b \otimes (1 - A_i),
\end{aligned}
\label{eq:att}
\end{equation}
where $\otimes$ denotes element-wise product.
In this way, both the image decoder and the attention decoder can interact with each other and ultimately produce better results.

\subsection{Model Training}
\noindent \textbf{Appearance and Shape Discriminators.} 
We adopt two discriminators for adversarial training.
Specifically, we feed image-image pair ($I_a$, $I_b$) and ($I_a$, $I_b^{'}$) into the appearance discriminator $D_a$ to ensure appearance consistency.
Meanwhile, we feed pose-image pair ($P_b$, $I_b$) and ($P_b$, $I_b^{'}$) into the shape discriminator $D_s$ for shape consistency.
Both discriminators (\emph{i.e.}, $D_a$ and $D_s$), and the proposed graph generator $G$ are trained in an end-to-end way, aiming to enjoy mutual benefits from each other in a joint framework.

\noindent \textbf{Optimization Objectives.} We follow \cite{zhu2019progressive,tang2020xinggan} and use the adversarial loss $\mathcal{L}_{gan}$, the pixel-wise $L1$ loss $\mathcal{L}_{l1}$ and the perceptual loss $\mathcal{L}_{per}$ as our optimization objectives,
\begin{equation}
\begin{aligned}
\mathcal{L}_{full} = \lambda_{gan} \mathcal{L}_{gan} + \lambda_{l1} \mathcal{L}_{l1} + \lambda_{per} \mathcal{L}_{per},
\end{aligned}
\label{eq:loss}
\end{equation}
where $\lambda_{gan}$, $\lambda_{l1}$ and $\lambda_{per}$ control the relative importance of the three objectives.
For the perception loss, we follow \cite{zhu2019progressive,tang2020xinggan} and use the $Conv1\_2$ layer.

\noindent \textbf{Implementation Details.}
In our experiments, we follow previous work~\cite{zhu2019progressive,tang2020xinggan} and represent the source pose $P_a$ and the target pose $P_b$ as two 18-channel heat maps that encode the locations of 18 joints of a human body.
Adam optimizer \cite{kingma2014adam} is employed to learn the proposed BiGraphGAN for around 90K iterations with $\beta_1{=}0.5$ and $\beta_2{=}0.999$.

In preliminary experiments, we found that as $T$ increases, the performance is getting better and better. When $T$ is equal to 9, the proposed model achieves the best results, and then the performance begins to decline. Thus we set $T{=}9$ in the proposed graph generator.
Moreover, $\lambda_{gan}$, $\lambda_{l1}$, $\lambda_{per}$ in Eq.~\eqref{eq:loss}, and the number of feature map channels $C$ are set to 5, 10, 10, and 128, respectively.
The proposed BiGraphGAN is implemented in PyTorch~\cite{paszke2019pytorch}. 

\section{Experiments}

\begin{table*}[!t]
	\centering
	\caption{Quantitative comparison of different methods on Market-1501 and DeepFashion. For all metrics, higher is better. ($\ast$) denotes the results tested on our testing set.}
	 	\resizebox{1\linewidth}{!}{%
	\begin{tabular}{lcccccccc} \toprule
		\multirow{2}{*}{Method}  & \multicolumn{5}{c}{Market-1501} & \multicolumn{3}{c}{DeepFashion} \\ \cmidrule(lr){2-6} \cmidrule(lr){7-9} 
		& SSIM & IS   & Mask-SSIM & Mask-IS  & PCKh  & SSIM  & IS  & PCKh      \\ \hline	
		PG2~\cite{ma2017pose}                                        & 0.253 & 3.460 & 0.792 & 3.435   & - & 0.762 & 3.090  & - \\
		DPIG~\cite{ma2018disentangled}                           & 0.099 & 3.483 & 0.614 & 3.491   & - & 0.614 & 3.228    & - \\
		Deform~\cite{siarohin2018deformable}              & 0.290 & 3.185 & 0.805 & 3.502   & - & 0.756 & 3.439    & -\\ 
		C2GAN~\cite{tang2019cycle}                                & 0.282 & 3.349 & 0.811 & 3.510   & - & -     & -            & -\\ 
		BTF~\cite{albahar2019guided}                               & -     & -     & -     & -       & - & 0.767 & 3.220                   & -\\ \hline
		PG2$^\ast$~\cite{ma2017pose}                             & 0.261 & \textbf{3.495} & 0.782 & 3.367   &0.73 & 0.773 & 3.163  & 0.89 \\ 
		Deform$^\ast$~\cite{siarohin2018deformable}    & 0.291 & 3.230 & 0.807 & 3.502   & \textbf{0.94} & 0.760 & 3.362 & 0.94 \\ 
		VUnet$^\ast$~\cite{esser2018variational}              & 0.266 & 2.965 & 0.793 & 3.549   & 0.92 & 0.763 & \textbf{3.440} & 0.93 \\
		PATN$^\ast$~\cite{zhu2019progressive}  & 0.311 & 3.323 & 0.811 & \textbf{3.773}   & \textbf{0.94} & 0.773 & 3.209 & 0.96 \\  
		BiGraphGAN  & \textbf{0.325} &  3.329 & \textbf{0.818} & 3.695 & \textbf{0.94} & \textbf{0.778} & 3.430 & \textbf{0.97}  \\ \hline	
		Real Data                                                               & 1.000 & 3.890 & 1.000 & 3.706   & 1.00 & 1.000 & 4.053 & 1.00 \\	
		\bottomrule	
	\end{tabular}}
	\label{tab:pose_reuslts}
\end{table*}

\begin{figure}[!t]
	\centering
	\includegraphics[width=1\linewidth]{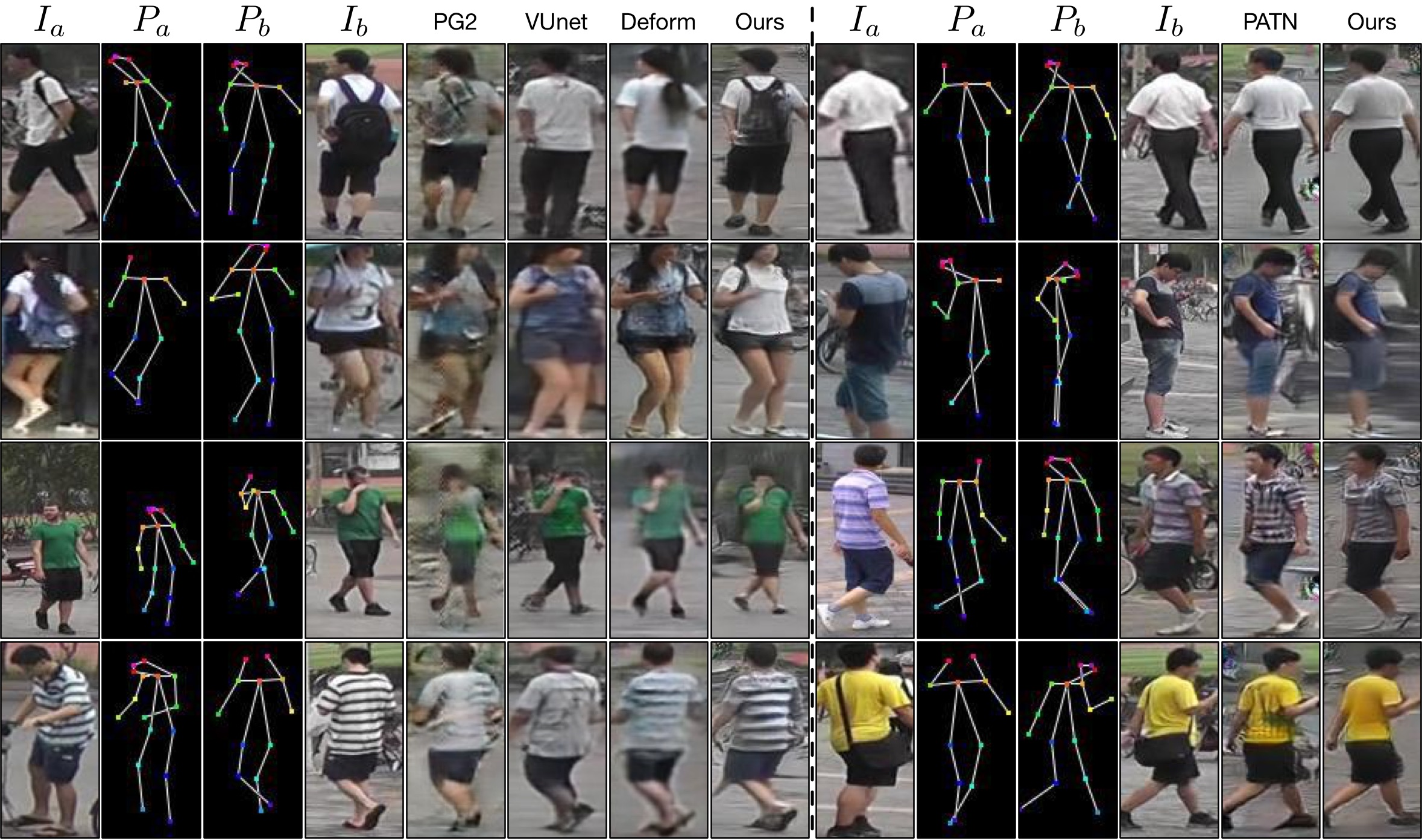}
	\caption{Qualitative comparisons of different methods on Market-1501.}
	\label{fig:mark_results}
\end{figure}

\noindent \textbf{Datasets.} 
We follow previous works \cite{ma2017pose,siarohin2018deformable,zhu2019progressive} and conduct extensive experiments on two public datasets, \emph{i.e.}, Market-1501 \cite{zheng2015scalable} and DeepFashion \cite{liu2016deepfashion}.
Specifically, we adopt the train/test split used in \cite{zhu2019progressive,tang2020xinggan} for a fair comparison.
In addition, images are resized to $128 {\times} 64$ and $256 {\times} 256$ on Market-1501 and DeepFashion, respectively.
 
\noindent \textbf{Evaluation Metrics.}
We follow \cite{ma2017pose,siarohin2018deformable,zhu2019progressive} and employ Inception score (IS) \cite{salimans2016improved}, Structure Similarity (SSIM) \cite{wang2004image} and their masked versions (\emph{i.e.}, Mask-IS and Mask-SSIM) as our evaluation metrics to quantitatively measure the quality of the generated images by different approaches.
Moreover, we employ the PCKh score proposed in \cite{zhu2019progressive} to explicitly evaluate the shape consistency of the generated person images.



\subsection{State-of-the-Art Comparisons}

\noindent \textbf{Quantitative Comparisons.}
We compare the proposed BiGraphGAN with several leading person image synthesis methods, \emph{i.e.}, PG2~\cite{ma2017pose}, DPIG~\cite{ma2018disentangled}, Deform~\cite{siarohin2018deformable,siarohin2019appearance}, C2GAN~\cite{tang2019cycle}, BTF~\cite{albahar2019guided}, VUnet~\cite{esser2018variational}, and PATN~\cite{zhu2019progressive}.
Quantitative comparison results are shown in Table~\ref{tab:pose_reuslts}, we can see that the proposed method achieves the best results on most metrics such as SSIM, Mask-SSIM and PCKh on Market-1501, and SSIM and PCKh on DeepFashion.
For other metrics such as IS, the proposed method still achieves better results than the most related model PATN on both datasets.
These results validate the effectiveness of our method.

\begin{figure}[!t]
	\centering
	\includegraphics[width=1\linewidth]{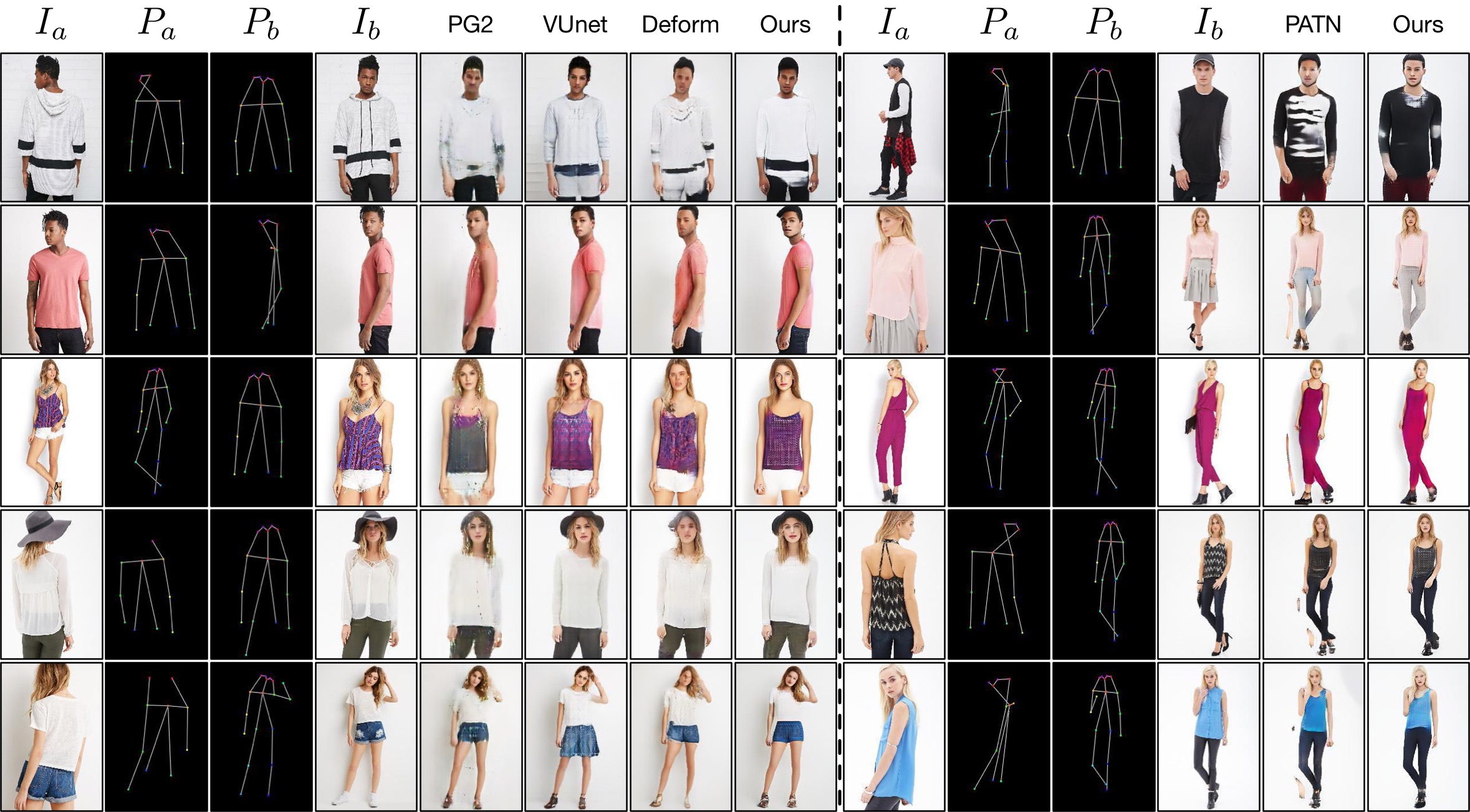}
	\caption{Qualitative comparisons of different methods on DeepFashion.}
	\label{fig:fashion_results}
\end{figure}

\noindent \textbf{Qualitative Comparisons.}
We also provide visualization comparison results on both datasets in Fig.~\ref{fig:mark_results} and \ref{fig:fashion_results}.
As shown in the left of both figures, the proposed BiGraphGAN generates remarkably better results than PG2~\cite{ma2017pose}, VUnet~\cite{esser2018variational} and Deform~\cite{siarohin2018deformable} on both datasets.
To further evaluate the effectiveness of the proposed method, we compare the proposed BiGraphGAN with the most state-of-the-art model, \emph{i.e.}, PATN~\cite{zhu2019progressive}, in the right of both figures.
We still observe that our proposed BiGraphGAN generates more clear and visually plausible person images than PATN on both datasets.

\begin{table}[!t]
	\centering
	\caption{Quantitative comparison of user study (\%) on Market-1501 and DeepFashion. `R2G' and `G2R' represent the percentage of real images rated as fake \emph{w.r.t.}~all real images, and the percentage of generated images rated as real \emph{w.r.t.} all generated images, respectively.}
	\begin{tabular}{lccccccc} \toprule
		\multirow{2}{*}{Method}  & \multicolumn{2}{c}{Market-1501} & \multicolumn{2}{c}{DeepFashion} \\ \cmidrule(lr){2-3} \cmidrule(lr){4-5} 
		& R2G & G2R & R2G & G2R    \\ \hline	
		PG2~\cite{ma2017pose}                              & 11.20  & 5.50    & 9.20   & 14.90 \\
		Deform~\cite{siarohin2018deformable}    & 22.67 & 50.24  & 12.42 & 24.61 \\ 
		C2GAN~\cite{tang2019cycle}                      & 23.20 & 46.70  & -     & -     \\
		PATN~\cite{zhu2019progressive}   & 32.23 & 63.47  & 19.14 & 31.78 \\  
		BiGraphGAN                                       & \textbf{35.76} & \textbf{65.91}  & \textbf{22.39} & \textbf{34.16}  \\	
		\bottomrule	
	\end{tabular}
	\label{tab:pose_ruser}
\end{table}

\noindent \textbf{User Study.}
We also follow \cite{ma2017pose,siarohin2018deformable,zhu2019progressive} and conduct a user study to evaluate the quality of the generated images.
Specifically, we follow the evaluation protocol used in \cite{zhu2019progressive} for a fair comparison.
Comparison results of different methods are shown in Table~\ref{tab:pose_ruser}, we can see that the proposed method achieves the best results on all metrics, which further validates that the generated images by the proposed BiGraphGAN are more photo-realistic.

\subsection{Ablation Study}
\begin{figure}[!t]
	\centering
	\includegraphics[width=1\linewidth]{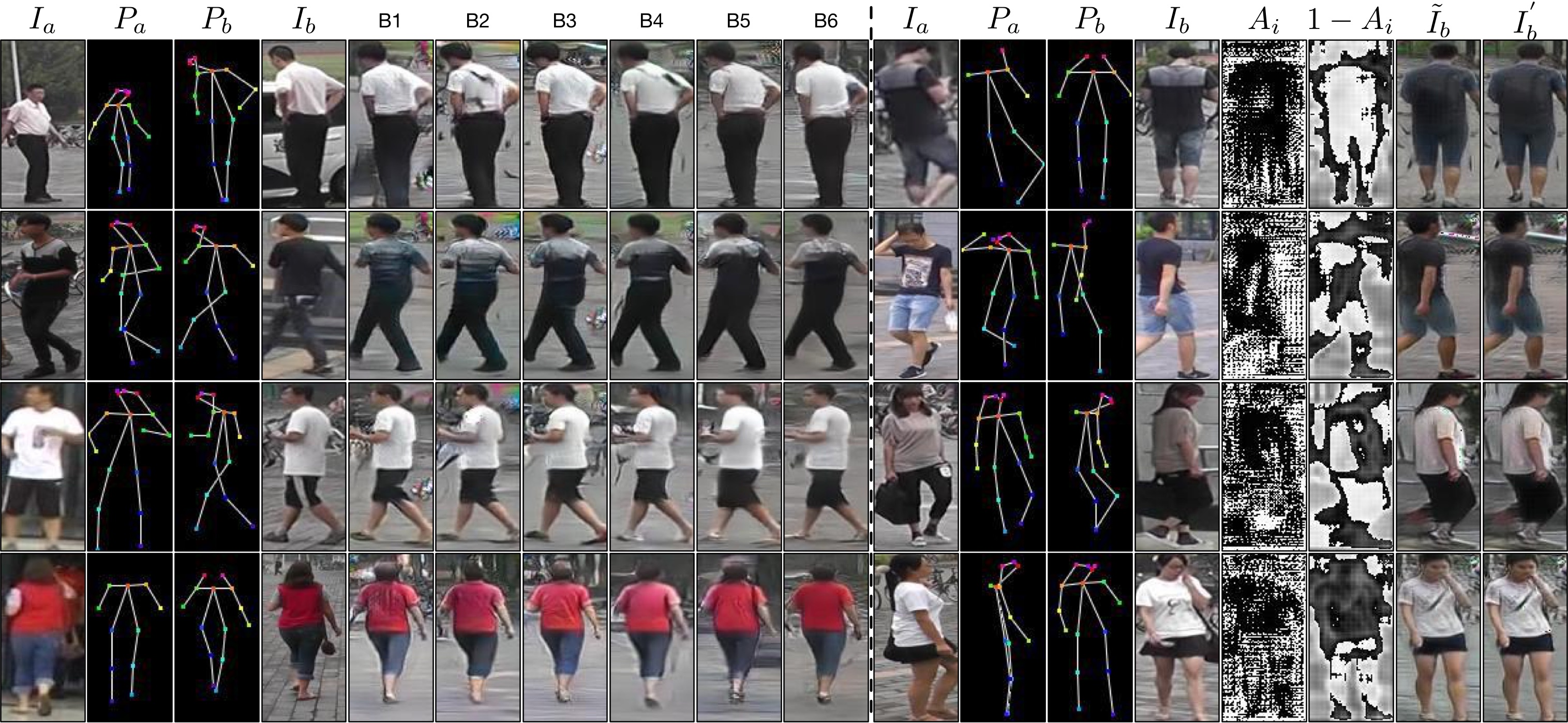}
	\caption{(\textit{left}) Qualitative comparisons of ablation study on Market-1501. (\textit{right}) Visualization of the learned attention masks and intermediate results.}
	\label{fig:ablation}
\end{figure}

\noindent \textbf{Baselines of BiGraphGAN.}
We perform extensive ablation studies to validate the effectiveness of each component of the proposed BiGraphGAN on Market-1501. 
The proposed BiGraphGAN has 6 baselines (\emph{i.e.}, B1, B2, B3, B4, B5, B6) as shown in Table \ref{tab:ablation} and Fig.~\ref{fig:ablation}(\textit{left}).
B1 is our baseline.  
B2 uses the proposed B2A branch for modeling the crossing relations from the target pose to the source pose.
B3 adopts the proposed A2B branch to model the crossing relations from the source pose to the target pose.
B4 uses the combination of both A2B and B2A branches to model the crossing relations between the source pose and the target pose.
Note that both GCNs in B4 are sharing the parameters.
B5 employs a non-sharing strategy between the two GCNs to model the crossing relations.
B6 employs the proposed AIF module to make the graph generator attentively select which part is more useful for generating the final person image.

\begin{table}[!t]
	\centering
	\caption{Ablation study of the proposed BiGraphGAN on Market-1501. For both metrics, higher is better.}
	\begin{tabular}{lcc} \toprule
		Baselines of BiGraphGAN & SSIM $\uparrow$ & Mask-SSIM $\uparrow$    \\ \midrule
		B1: Our Baseline & 0.305 & 0.804 \\ 
		B2: B1 + B2A     & 0.310 & 0.809 \\
		B3: B1 + A2B     & 0.310 & 0.808 \\
		B4: B1 + A2B + B2A (Sharing) & 0.322 & 0.813 \\
		B5: B1 + A2B + B2A (Non-Sharing) & 0.324 & 0.813 \\
		B6: B5 + AIF & \textbf{0.325} & \textbf{0.818} \\
		\bottomrule	
	\end{tabular}
	\label{tab:ablation}
\end{table}

\noindent \textbf{Ablation Analysis.}
The results of the ablation study are shown in Table \ref{tab:ablation} and Fig.~\ref{fig:ablation}(\textit{left}).
We observe that both B2 and B3 achieve significantly better results than B1, which proves our initial motivation that modeling the crossing relations between the source pose and the target pose in a bipartite graph will boost the generation performance.
In addition, we see that B4 performs better than B2 and B3, demonstrating the effectiveness of modeling the symmetric relations between the source and target poses.
B5 achieves better results than B4, which means that two GCNs are constructed separately to model the symmetric relations will improve the generation performance in the joint network.
B6 is better than B5, which clearly proves the effectiveness of the proposed attention-based image fusion strategy.

Moreover, we show several examples of the learned attention masks and intermediate results in Fig.~\ref{fig:ablation}(\textit{right})
We can see that the proposed module attentively selects useful content from both the input image and intermediate result to generate the final result, thus verifying our design motivation.

\section{Conclusions}
In this paper, we propose a novel Bipartite Graph Reasoning GAN (BiGraphGAN) framework for the challenging person image generation task.
We introduce two novel blocks, \emph{i.e.}, Bipartite Graph Reasoning (BGR) block and Interaction-and-Aggregation (IA) block.
The first is employed to model the crossing long-range relations between the source pose and the target pose in a bipartite graph.
The second block is used to interactively enhance both person's shape and appearance features.
Extensive experiments of both human judgments and automatic evaluation demonstrate that the proposed BiGraphGAN achieves remarkably better performance than the state-of-the-art approaches.

\section*{Acknowledgment}
This work has been partially supported by the Italy-China collaboration project TALENT, the Royal Academy of Engineering under the Research Chair and Senior Research Fellowships scheme, EPSRC/MURI grant EP/N019474/1 and FiveAI.

\clearpage
\bibliography{egbib}

\end{document}